\documentclass[letterpaper, 10 pt, conference]{ieeeconf}  

\IEEEoverridecommandlockouts                              

\usepackage{graphics} 
\usepackage{epsfig} 
\usepackage{mathptmx} 
\usepackage{times} 
\usepackage{amsmath} 
\usepackage{amssymb}  
\usepackage{xcolor}

\let\labelindent\relax
\usepackage[bookmarks=true]{hyperref}
\hypersetup{
  colorlinks   = true, 
  urlcolor     = blue, 
  linkcolor    = blue, 
  citecolor   = red 
}
\usepackage{cleveref} 

\usepackage{multicol}
\usepackage[shortlabels]{enumitem}
\usepackage{subcaption}

\usepackage[
    style=ieee,
    sorting=none
]{biblatex}
\crefname{figure}{Fig.}{Fig.}
\Crefname{figure}{Figure}{Figures}
\Crefname{equation}{Equation}{Equations}
\crefname{theorem}{Theorem}{Theorems}
\crefname{section}{Section}{Sections}
\crefname{hypothesis}{Hypothesis}{Hypotheses}
\crefname{proposition}{Proposition}{Propositions}
\crefname{algorithm}{Algorithm}{Algorithm}

\usepackage{fancyhdr}
\fancypagestyle{withfooter}{
  
  \fancyfoot[C]{\footnotesize Accepted to the IEEE ICRA Workshop on Field Robotics 2024}
}

\title{\LARGE \bf
Modular, Resilient, and Scalable System Design Approaches - Lessons learned in the years after DARPA Subterranean Challenge
}

\author{Prasanna Sriganesh, James Maier, Adam Johnson, Burhanuddin Shirose, Rohan Chandrasekar, \\ Charles Noren, Joshua Spisak, Ryan Darnley, Bhaskar Vundurthy and Matthew Travers 
\thanks{ All the authors are affiliated with Carnegie Mellon University. Prasanna Sriganesh (\texttt{pkettava@andrew.cmu.edu}) is the corresponding author.  }%
}
\begin{document}

\maketitle
\thispagestyle{withfooter}
\pagestyle{withfooter}

\begin{abstract}



Field robotics applications, such as search and rescue, involve robots operating in large, unknown areas. These environments present unique challenges that compound the difficulties faced by a robot operator. The use of multi-robot teams, assisted by carefully designed autonomy, help reduce operator workload and allow the operator to effectively coordinate robot capabilities. In this work, we present a system architecture designed to optimize both robot autonomy and the operator experience in multi-robot scenarios. Drawing on lessons learned from our team's participation in the DARPA SubT Challenge, our architecture emphasizes modularity and interoperability. We empower the operator by allowing for adjustable levels of autonomy (``sliding mode autonomy"). We enhance the operator experience by using intuitive, adaptive interfaces that suggest context-aware actions to simplify control. Finally, we describe how the proposed architecture enables streamlined development of new capabilities for effective deployment of robot autonomy in the field.

\end{abstract}


\section{Introduction}

Field robotics encompasses a wide range of applications in which robots must operate in large, unknown and often harsh environments. One such application is a search-and-rescue mission, where an operator deploys a robot to map and explore hazardous environments. These missions have traditionally relied on teleoperation, for instance, robots were deployed at the Fukushima Daiichi nuclear reactor disaster~\cite{fukushimadaiichi} or the snake robots used in the 2017 Mexico earthquakes~\cite{whitman2018snake}. However, direct teleoperation places a high cognitive burden on the operator, especially in complex environments with limited communications \cite{chen2007human}. Robot autonomy becomes a crucial tool in these scenarios, allowing the operator to focus on high-level mission goals. 


Robot autonomy has received promising improvements to the performance driven by recent advances in robot perception, localization, planning, and artificial intelligence (AI) systems. The 2021 DARPA Subterranean (SubT) Challenge~\cite{darpa_tim} provided a compelling showcase for these advancements. SubT involved a single human operator deploying a team of heterogeneous robots for coordinated mapping and exploration in unknown subterranean scenarios. While the challenge highlighted the potential of robot autonomy, it exposed the teams to problems of real-world deployment. The analysis of DARPA SubT Challenge results~\cite{darpa_tim} identified several key challenges, particularly: operator cognitive overload from managing multiple robots, robot attrition, and ensuring interoperability within a heterogeneous team. In order to address the identified challenges, we posit that developing a holistic systems-level design (e.g., from individual robots to the operator experience/UI) is a crucial step for the effective deployment of autonomous systems in the field


We present a system architecture designed to optimize both robotic autonomy and the operator experience within the context of multi-robot deployments. We draw from our group's experience as part of Team Explorer in SubT and outline the lessons learned from the field. These lessons drive core elements of our design philosophy for the presented architecture, namely: modularity and adaptive autonomy. Designing a system with a focus on modularity enables the rapid configuration and deployment of heterogeneous robotic systems. Coupling this design focus with tools for adaptive autonomy, which includes both operator-adjustable levels of autonomy (for example, switching between assisted teleoperation and waypoint navigation) and behavior-tree-based intuitive user-interfaces, yields a system where operator workload does not scale as poorly with additional robots.  Finally, we outline how different components of the architecture work in tandem to enable easier development of future multi-robot capabilities without the need for extensive reconfiguration.

\section{Related Work}

In this section, we first describe relevant work and insights from field robotics research, with a focus on search-and-rescue and subterranean environments. We next present the different approaches used by the participating teams in the DARPA SubT challenge that highlight lessons learned in the deployment of multi-robot autonomy.

Autonomous exploration in subterranean environments has been a significant focus in field robotics. These include works involving mapping of mine environments~\cite{baker2004campaign}~\cite{thrun2004autonomous}, disaster response~\cite{fukushimadaiichi}~\cite{whitman2018snake}~\cite{nagatani2013emergency} and planetary exploration~\cite{agha2019robotic}. Recent advances has focused on improving specific aspects of subterranean robotics such as perception, localization and navigation. Robots operating in these environments place constraints on compute available, and demand development of lightweight architecture for object detection~\cite{liu2016ssd}~\cite{szegedy2016rethinking}. The degraded nature of the environment have led to development  of localization algorithms that can fuse information from multiple sources ~\cite{zhao2021super}~\cite{palieri2021corrections}. Novel algorithms have been developed for long-range exploration and planning for such environments~\cite{yang2022far}~\cite{bouman2020autonomous}. 

One of the biggest challenges presented by search-and-rescue scenarios~\cite{delmerico2019current}~\cite{murphy2016disaster} is the low-bandwidth and limited communication range. Robots are required to built their own networks, and exchange data only when necessary. This has led to algorithms to enable multi-robot exploration with limited communications~\cite{amigoni2017multirobot}. There have also been solutions to ensure that the robots maintain persistent communications by monitoring the network and dropping repeater nodes when necessary~\cite{tatum2020}. The DARPA Subterranean challenge brings all these various advancements together and exposes the challenges present in deployment.

In the analysis of the 2021 DARPA SubT Final Event by Dr. Timothy Chung \cite{darpa_tim}, the authors highlight the inevitability of robot attrition in real-world deployments. They emphasize on the importance of streamlined system integration to enable fast team reconfiguration. The authors also state that robot heterogeneity was crucial to team performance, and was partially enabled by the increasing availability of various mobile robot platforms. Finally, they underscore the importance of the human-robot interface as ``the human teammate is often the critical performance limiter''.  While a comprehensive survey of DARPA SubT technical approaches warrants a separate work, here we focus on key details from Team Explorer and other relevant teams that address these critical system-level insights.

\textbf{Team Explorer in the DARPA SubT Challenge:} Team Explorer's approach to SubT \cite{Travers2022} comprised of heterogeneous robot team controlled by a single operator. They emphasised the ``sliding mode autonomy framework"~\cite{dias2008sliding}~\cite{heger2006sliding}, that allowed the human operator to adjust the level of autonomy for different tasks. This framework enabled different robot operational modes with varying levels of autonomy (e.g., exploration, go-to-waypoint, manual etc). A behavior tree \cite{bt_survey} was used to arbitrate different behaviors on the robot. 

Team Explorer mitigated issues with high operator cognitive load by using a GUI which was highly tailored to the format of the SubT competition, while also employing human aids to the single operator who would remind the operator to complete certain tasks, thus letting the operator focus solely on operating robots.  Additionally, robot capabilities were defined through module-specific configuration files, enabling heterogeneity in the fleet. However, this approach limited on-the-fly reconfiguration, highlighting the need for enhanced interoperability in future systems.

\textbf{Other teams in the DARPA SubT Challenge:} The work described by other DARPA Subterranean Challenge performer teams provide great insights to address challenges in human-robot interfacing, and overall system resiliency. Team CoSTAR~\cite{costar}, describe a copilot assistant that manages the interface between the human operator and the autonomous system. This copilot assistant can create and schedule tasks to both the human as well as the robots by treating the operator and the robots as a single team and helps reduce the cognitive load on the operator.

Other approaches to human-robot collaboration include the use of finite state automata for decision-making, as seen in Team CSIRO Data61's work~\cite{csiro}. They employed a market-based task allocation approach to coordinate the actions of different robots in their team. The authors note that tuning the rewards for task allocation was a challenge, and needed to be tuned for various operational environments. Additionally, they also described the need for incorporating manual tasks to incorporate operator guidance into this framework. This motivates the need for adjustable levels of autonomy, to avoid conflicts between the operator's intended actions and robot's decision making.

The CTU-CRAS-NORLAB team~\cite{norlab}~\cite{norlab_final} acknowledge that component-level failures are inevitable, and is a result of working with cutting-edge hardware which is operating at its limits.  In order to address this, they design their stack to continually log system state and automatically restart components from the last known safe state when a failure is detected. This inspires us to carefully craft information flow in the system to ensure restarts do not cause conflicting commands to the robot.



\section{Lessons Learned}

Our experiences, combined with the analysis of prior work described in the previous section, have highlighted five crucial lessons learned for the effective deployment of robot autonomy in the field. They are described below.
\begin{enumerate}[start=1,label={\textcolor{blue}{(\bfseries L\arabic*)}}]
    \item \textbf{Operator-Adjustable Autonomy:} Ensuring the operator has the ability to adjust the level of robot autonomy is crucial for flexible and adaptable operation \label{l1:sliding}
    \item \textbf{System Interoperability:} Seamless interoperability between heterogeneous robots is essential. This allows for dynamic robot deployment depending on environmental factors \label{l2:interop}
    \item  \textbf{Centralized Control Flow:} Careful management of system-wide control flow is necessary to prevent conflicting commands and ensure coordinated robot behavior \label{l3:controlflow}
    \item \textbf{Adaptive User Interface:} An interface that dynamically adjusts to present only valid actions based on context can significantly reduce the operator's cognitive workload \label{l4:adapiface}
    \item \textbf{Extensible System Design:} Prioritize modularity and flexibility during system integration to accommodate future changes and the addition of new capabilities \label{l5:easydev}
\end{enumerate}
In the next section, we described our developed architecture that is designed with these learned lessons in mind
\begin{figure*}[t!]
    \centering
    \includegraphics[width = 0.95\linewidth]{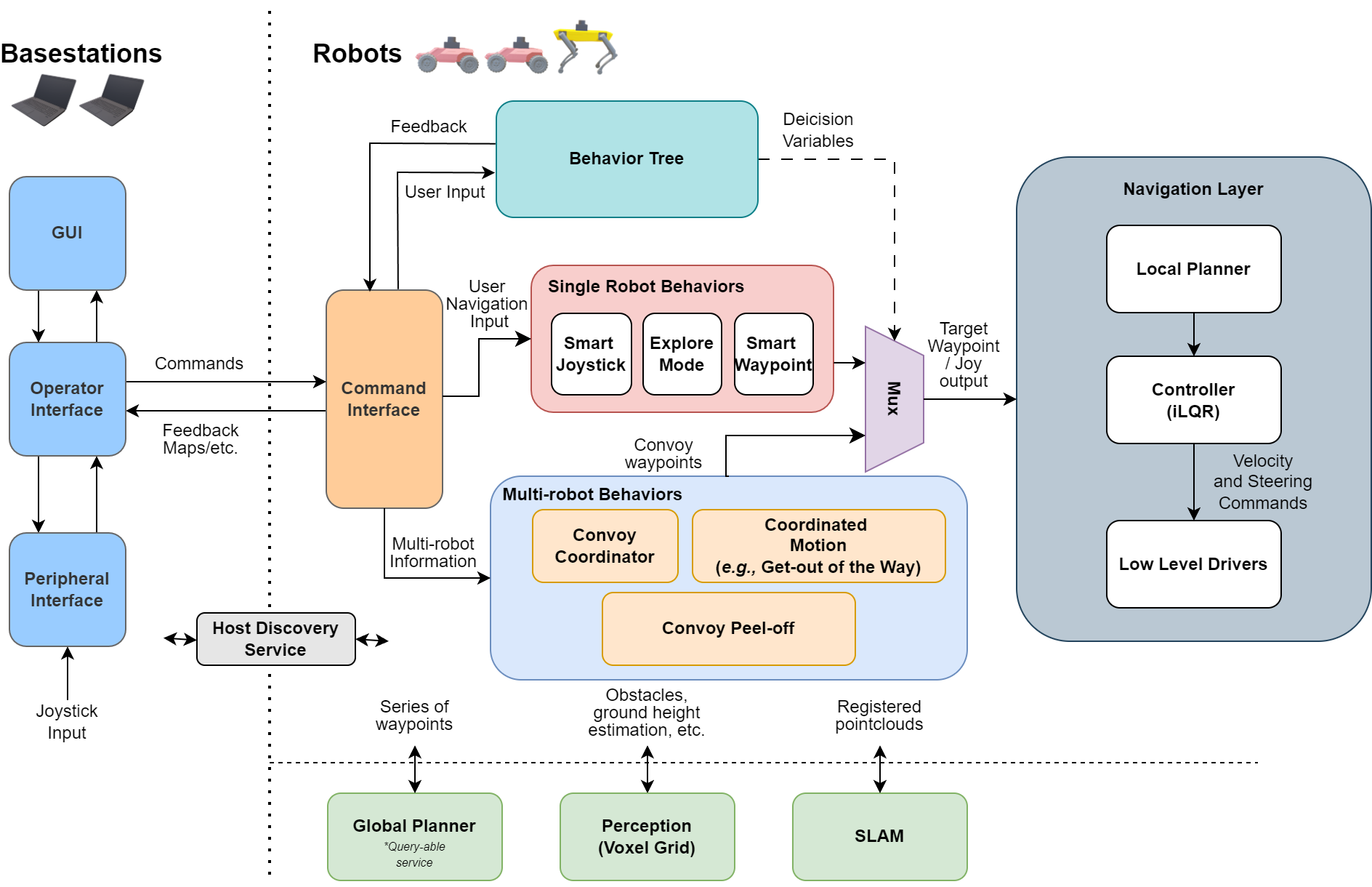}
    \caption{A overview of the proposed system architecture}
    \label{fig:archover}
    \vspace{-1em}
\end{figure*}

\section{Proposed Architecture}
This section describes the proposed architecture for a multi-robot system. Figure \ref{fig:archover} provides a visual overview. The architecture is designed to enable efficient coordination and control of a multi-robot system and consists of several modules serving different purposes. The architecture supports multiple operators (basestations) and multiple heterogeneous robots. We use Robot Operating System (ROS) \cite{quigley2009ros} for implementation and DDS for communication between the operator's basestation and the robots. This allows us to control information exchange and monitor communication bandwidth.

The architecture supports dynamically adding or removing robots in real-time. This is made possible by a host discovery service, which shares information about active agents (robots and basestations) on the network, and their capabilities (e.g., legged or wheeled robot). This is discussed further in section~\ref{HIT}.

We embrace the concept of sliding mode autonomy~\cite{dias2008sliding}~\cite{heger2006sliding} to ensure the operator can adjust the level of autonomy needed for a task while maintaining high-level control. This aligns with \ref{l1:sliding} and as a result, creates different operational modes (similar to Team Explorer's implementation \cite{Travers2022}) with increasing degrees of autonomy: 
\begin{itemize}
    \item \textbf{Full Manual Mode:} Operator commands directly control the robot 
    \item \textbf{Smart Joystick Mode:}  User-commanded direction is used in conjunction with a planner to avoid obstacles and navigate through narrow spaces
    \item \textbf{Waypoint Mode:} The robot navigates to a target location provided by the operator
    \item \textbf{Exploration Mode:}  The robot autonomously explores an area and operates with the highest level of autonomy
\end{itemize}
Translating the user inputs to navigation commands is handled in the single robot behavior block, while the behavior tree ensures that the robot can be set to a specific mode. This is discussed further in section \ref{sec:beh}.

In addition to these modes, our architecture enables easy addition and development of new modes for different coordinated behaviors~\cite{bagree2023distributed}. One such example is the convoy mode where multiple robots form a convoy formation. The operator controls the lead robot using the modes mentioned above while the other robots follow-the-leader while staying in formation, extending the sliding mode autonomy across multiple robots. This is further discussed in section~\ref{sec:multi}.  

\begin{figure*}[t!]
    \centering
    \begin{subfigure}[t]{.54\linewidth}
        \centering
         \includegraphics[width=0.98\linewidth]{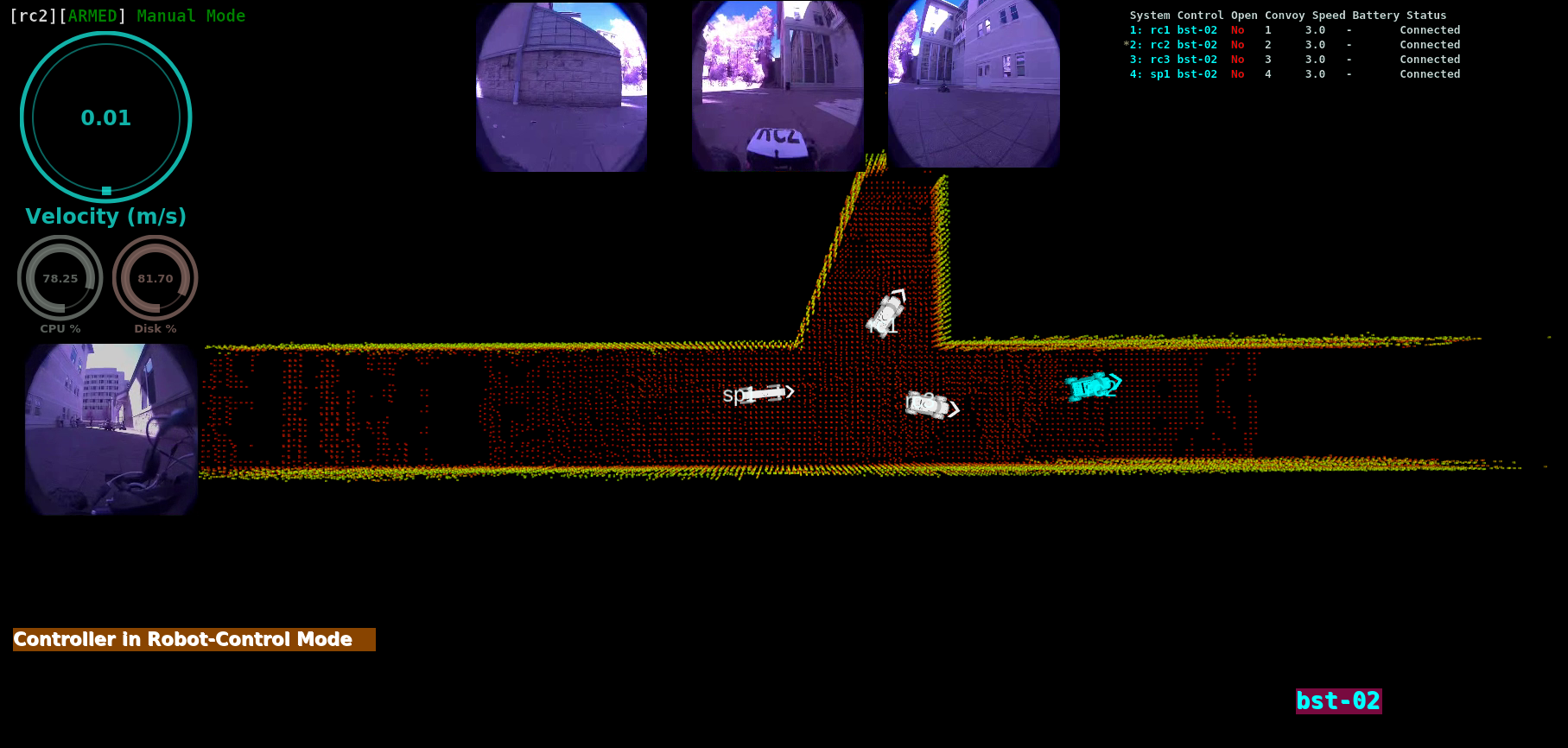}
        \caption{RViz visualization providing the operator with real-time status: map, robot locations, camera feeds of the active robot, its operational mode, and summary data for other robots on the top right}
        \label{fig:rviz}
    \end{subfigure}
    \begin{subfigure}[t]{.45\linewidth}
        \centering
         \includegraphics[width=0.99\linewidth]{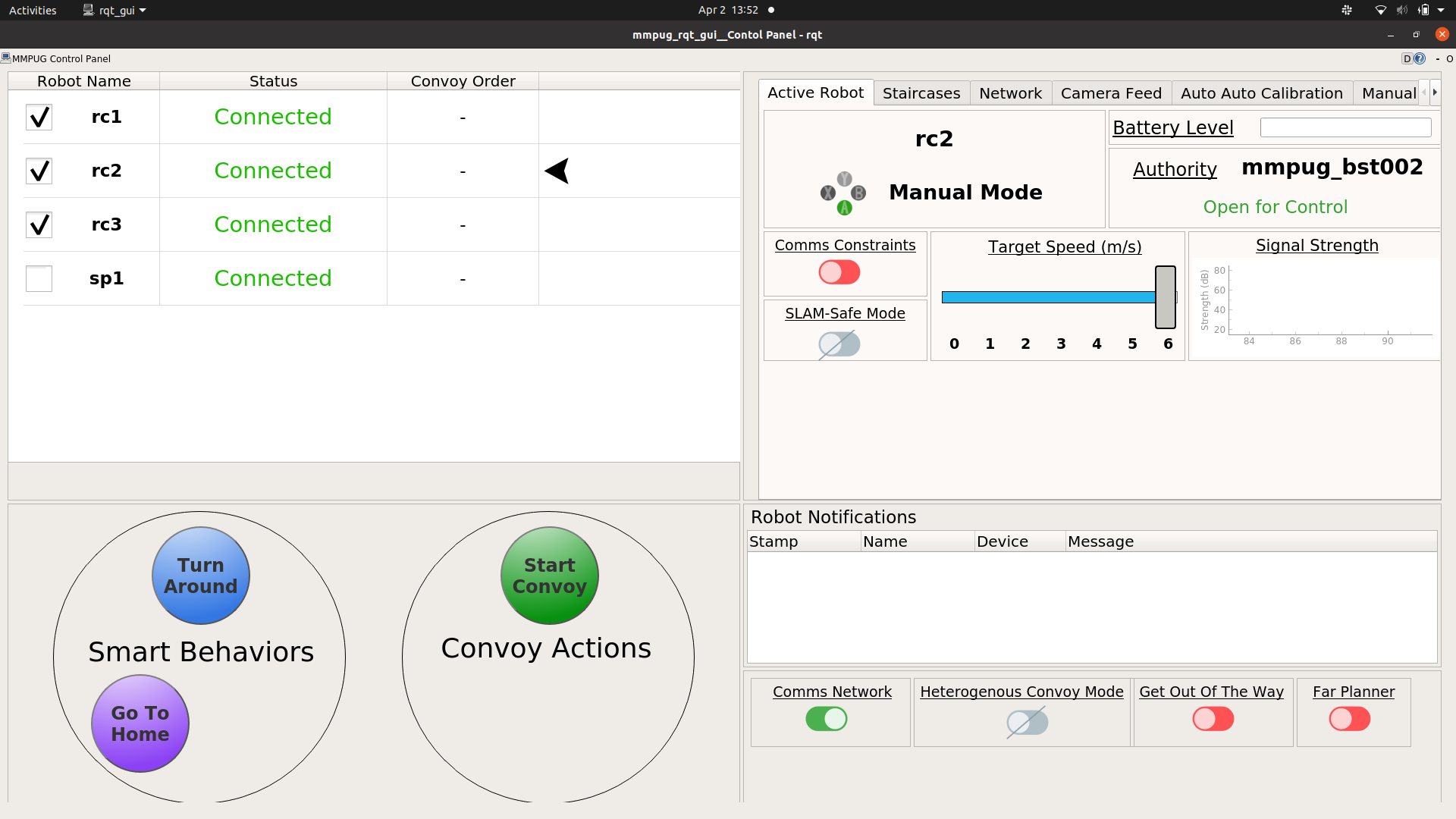}
        \caption{Touch-enabled GUI example: three robots selected (top left), active robot information (top right), valid actions (bottom left), and notifications/feature toggles (bottom right)}
        \label{fig:gui}
    \end{subfigure}
    \caption{Operator Interface: Example of RViz window, and the touch-enabled GUI}
    \vspace{-1.5em}
\end{figure*}

\subsection{Host discovery service}
\label{HIT}

To enable effective interoperability, our architecture requires knowledge about connected agents and their capabilities. We implement a host discovery service on each agent (robots and basestations) that provide this information. Each agent maintains a local server that stores identification information and its specific capabilities. Clients can query this server to obtain necessary information, allowing the selective launch of platform-specific modules. For example, a different controller and planner can automatically be launched in case of a quadruped.

Obtaining knowledge from other agents follows a ``request-response" communication model. The requesting agent queries the server of the responding agent, who replies with its identification and capabilities. To support discovery of an arbitrarily large number of agents, requests are sent in a multicast fashion.  Responding agents package the requested information and reply directly (unicast) to the requesting agent. This host discovery service optimizes communication on bandwidth-constrained networks while enabling discovery and transfer of robot-specific parameters. 

A published list of robots and basestations are provided over a ROS topic, ensuring system-wide awareness of active agents. We mandate that all applicable nodes must dynamically update their data streams based on this list. This host discovery process enables the seamless deployment of new robots and allows for better robot interoperability \ref{l2:interop}, a significant improvement over our previous work, which lacked this flexibility. 

\subsection{Command Interface} \label{sec:arch_cmdInt}
The command interface module serves as the component enabling multi-operator, multi-robot interaction. Its primary function is to receive all messages transmitted over the communications channel and determine whether they should be rejected or distributed to other modules within a robot's system. It also publishes robot telemetry data back to the basestation. The data sent to the basestation includes the robot state, the behavior tree state, downsampled map and camera streams. 

 Upon the host discovery service's updates, a list of the available devices is published throughout the network of robots and operators. The command interface monitors this list to track all available operators in the system. To ensure single-operator control at a time, the module keeps track of the current authorized basestation and grants the robot's control to any basestation upon request. Once authorized, only messages originating from the authorized basestation are forwarded to the rest of the stack. Messages from other basestations are discarded. 
 
The command interface also serves as the arbiter of control authority in the presence of multiple basestations. During operations, any basestation may place a request for control over a particular robot. Before the authority is assigned to a requesting basestation, the request is evaluated by the command interface to ensure that no other basestation currently holds a ``lock'' on the current robot. This ``lock'' designates that the vehicle is currently under the control authority of a different basestation and can only be overwritten when the basestation that is current authorized to command the robot releases the active lock.

The command interface ensures that there is only a single active interface point from different operators to a robot. This ensures that no conflicting commands are sent to the robot from different operators, addressing our lessons learned from \ref{l3:controlflow}. The use of the host discovery service alongside the command interface allows for clear interoperability with multiple operators \ref{l2:interop}.

\subsection{Behavior Tree and Mux} \label{sec:beh}
\begin{figure*}[t!]
    \centering
    \includegraphics[width=0.95\linewidth]{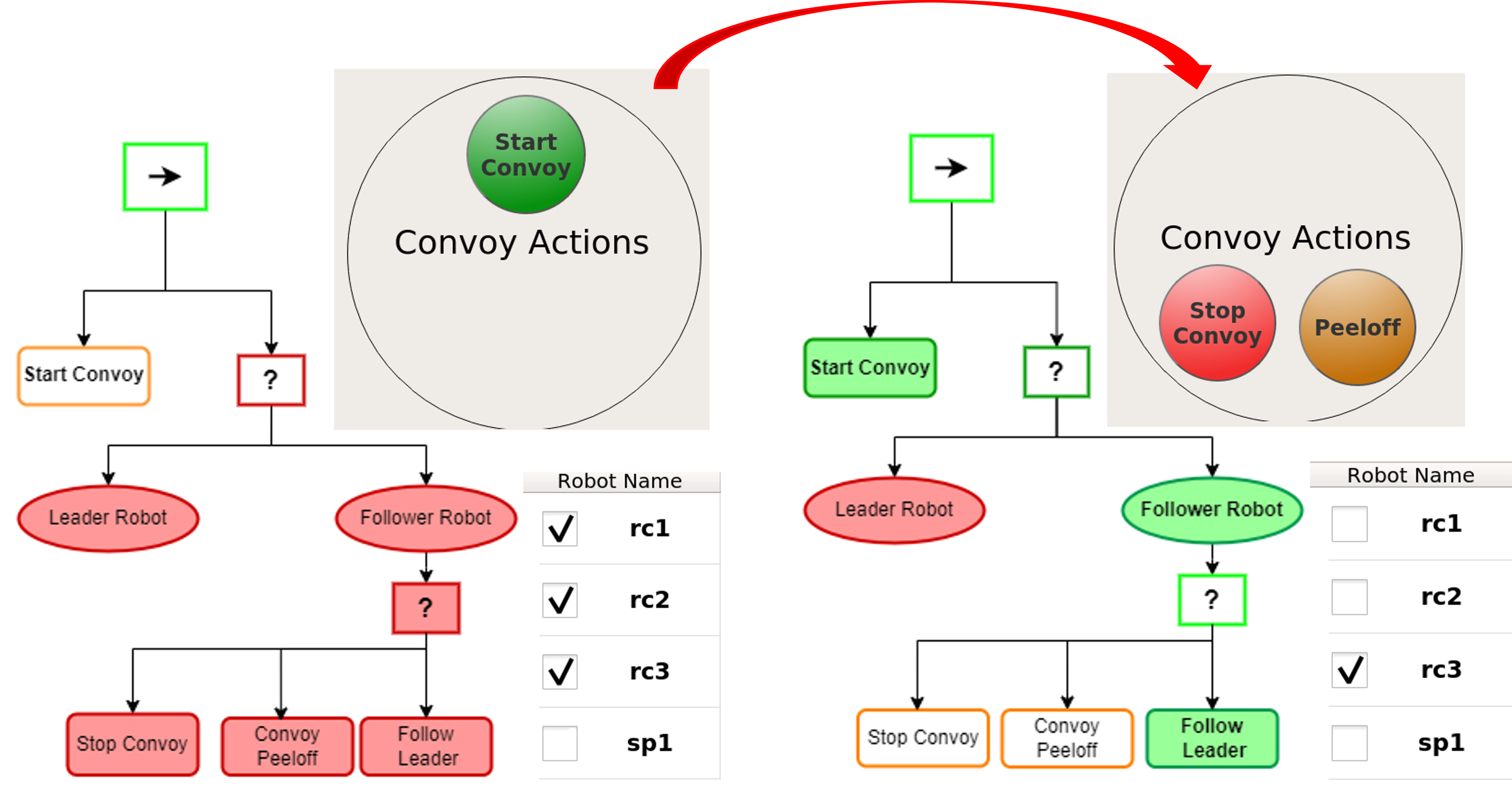}
    \caption{The left image shows the behavior subtree for the convoy with multiple robots selected, enabling the `start convoy' action. The right image demonstrates how the GUI dynamically updates based on the changes in the behavior subtree when a follower robot in an active convoy is selected, presenting actions like `peeloff' and `stop convoy'.}
    \label{fig:bt_gui}
    \vspace{-1em}
\end{figure*}

Behavior trees are central to our approach, managing complex system-mode transitions, enhancing modularity, and ensuring a clear structure \cite{colledanchise2018behavior}. Within our proposed architecture, the behavior tree acts as the central authority, determining the active mode on the robot. The behavior tree receives requests from the command interface, assesses their viability, switches the robot to the requested mode and enables that corresponding channel on the mux.

Each robot runs its own behavior tree, validating requests against predefined conditions. These conditions can be based on various things such as hardware availability, environmental constraints, or subsystem health or statuses. For instance, if the robot is not receiving joystick commands or if SLAM (Simultaneous Localization and Mapping) is not initialized, a `smart joystick mode' request will be discarded, as the `smart joystick mode' requires SLAM to function correctly. This structure of the behavior tree makes it easier and faster to define and implement new modes, along with their activation conditions \ref{l5:easydev}. 
 
Once the behavior tree determines the operational mode, it communicates this decision to the mux. The mux filters the outputs from different subsystems given the current robot mode specified by the behavior tree. This ensures that the navigation layer does not receive conflicting targets, promoting a centralized control flow in the system~\ref{l3:controlflow}. It also facilitates seamless integration of different subsystems~\ref{l5:easydev}.
 
The current behavior tree state (operational mode, authorized basestation, convoy order etc.) is sent back to the basestation as feedback. This helps in creating an adaptive interface that presents valid options to the user based on the operator's selection and the feedback from robot behavior trees. \ref{l4:adapiface}

\subsection{Operator Interface and GUI}
\begin{figure}[b!]
    \centering
    \includegraphics[width = 0.9\linewidth]{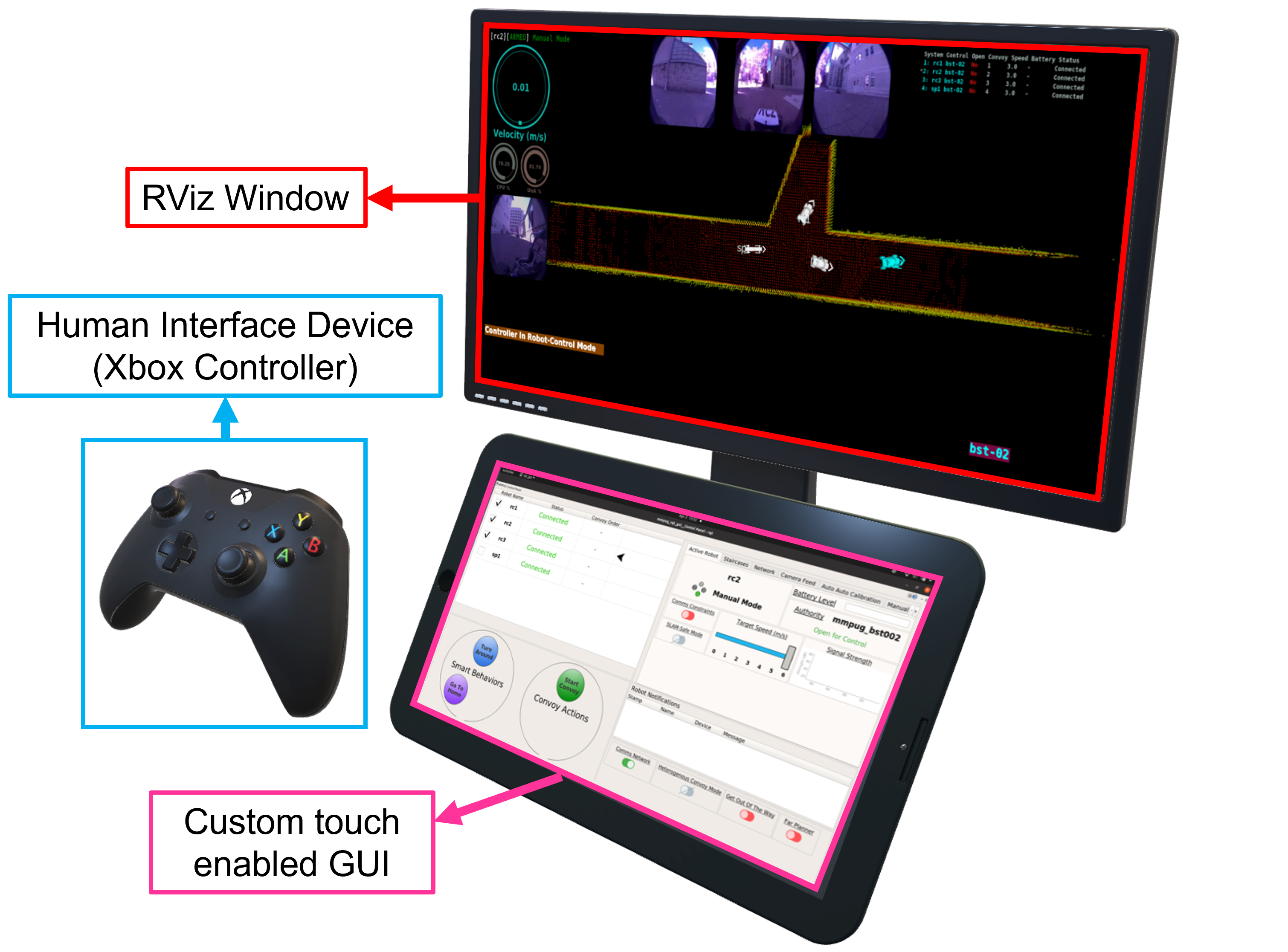}
    \caption{A pictorial representation of the operator setup showing the two screens}
    \label{fig:opsetup}
\end{figure}


The operator interface acts as the central hub for communication and control, providing the operator with real-time feedback from the connected robots. It features two screens:

\begin{itemize}
    \item \textbf{RViz Window:} Displays a 3D map, camera streams, robot modes, velocities, and overall system status (see Fig. \ref{fig:rviz}). This provides crucial situational awareness.
    \item \textbf{Custom GUI (Touch-Enabled):} Delivers subsystem feedback and enables rapid command actions through its touch-optimized design. (see Fig. \ref{fig:gui})
\end{itemize}
Figure \ref{fig:opsetup} shows the representation of the operator setup to control the robots.

Commands can be issued through either a physical human interface device (e.g., xbox controller) or the touch-enabled GUI on the basestation. The GUI dynamically updates based on the selected robots and their state feedback from the behavior tree. It presents only valid actions in real-time, guiding the operator's choices and minimizing errors. For example, the `start convoy' action becomes available only when multiple robots are selected, and  actions such as `stop convoy' and `peeloff' are displayed when a follower robot in an active convoy is selected (Fig. \ref{fig:bt_gui}).

The GUI also displays information about the active robot such as the mode of operation, the operator in control, and communication signal strength. The GUI features a notification panel for prioritized messages from the robot, such as control requests from other operators or subsystem failures. Additionally, the GUI has tabs for specialized information like staircase detection, network monitoring etc. These tabs are intentionally categorized and hidden, allowing the operator to access relevant details without being overwhelmed.

Our architecture acknowledges the reality of component-level failures. A dedicated tab displays the health status of individual nodes. This empowers the operator to choose between automatic or manual restarts. This custom GUI directly addresses the fourth lesson learned: the importance of adaptive user interfaces \ref{l4:adapiface}.

\begin{figure*}[t!]
    \includegraphics[width=0.99\linewidth]{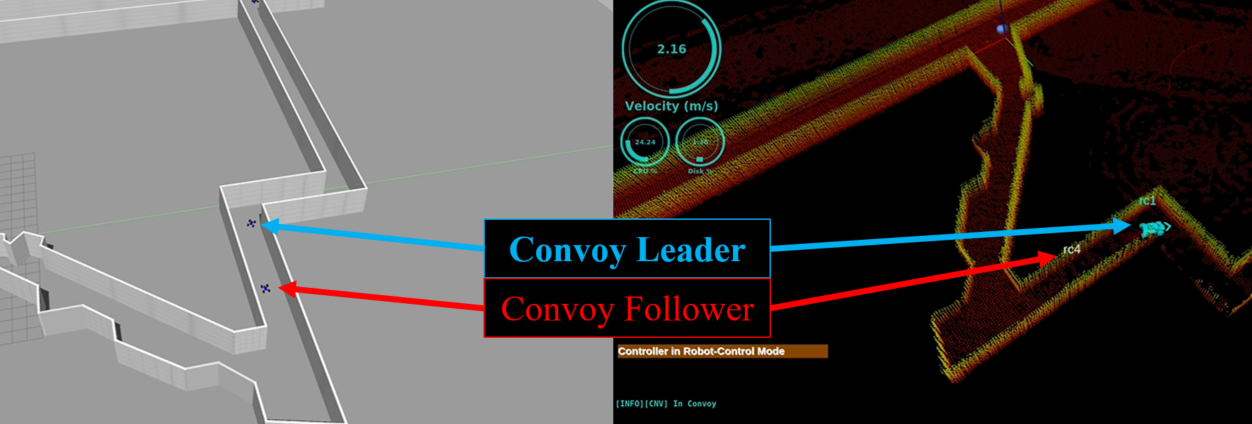}
    \caption{Simulation of two robots in convoy formation}
     \label{fig:convoy}
     \vspace{-1em}
\end{figure*}

\subsection{Navigation Layer}
The navigation layer serves as the component responsible for translating high-level operator commands into actionable directives for the robot's hardware. The navigation layer is typically made up of a local planner that compute a path towards the goal (e.g., trajectory library~\cite{zhang2020falco}), a controller to execute this plan (e.g., iLQR controller \cite{mayne1973differential}), and the low-level controllers (drivers) that translate these control actions to the actuators. By keeping these components separate, the architecture allows for flexibility and easier testing. One can swap out the planner or controller for different robots or experiment with new navigation algorithms without overhauling the entire system \ref{l5:easydev}.

\begin{figure}[b!]
    \centering
    \includegraphics[width = 0.9\linewidth]{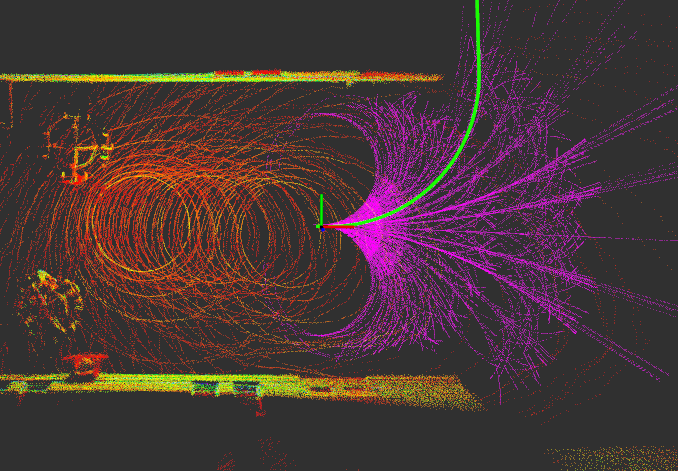}
    \caption{Robot employing a trajectory library based local planner. All feasible paths are shown in magenta, and selected path is shown in green}
    \label{fig:lplan}
\end{figure}

A single interface point to the navigation layer from the mux prevents unintended commands. This structure ensures that conflicting directives from the autonomy stack cannot be received by the navigation layer. The mux, controlled by the behavior tree, determines the true input source, promoting a centralized control flow and enhancing safety \ref{l4:adapiface}. Figure \ref{fig:lplan} shows how trajectory libraries are used to select a path for the robot controller to track.

\subsection{Advanced Multi-agent Coordination} \label{sec:multi}

The proposed architecture simplifies the development and implementation of coordinated multi-agent behaviors, contributing to reduced operator workload. The use of behavior trees and the mux as a single interface point provides a streamlined framework for implementing features, including multi-robot convoying and get-out-of-the-way maneuvers~\cite{bagree2023distributed,Toth2014,Nahavandi2022}.
 
In the convoy scenario, the operator directly controls the lead robot in either the smart joystick or waypoint mode. The follower robots autonomously maintain formation, tracking the lead robot's movement. This is achieved through the dedicated `convoy coordinator' block in the architecture (\ref{fig:archover}). This block handles information about all the robots in convoy and computes the target waypoints for the follower robots. These waypoints are then sent to the mux, and the behavior tree directs the mux to forward inputs from this block to the navigation layer. Figure \ref{fig:convoy} shows a simulated convoy with 2 robots.

Developing new capabilities is simplified with this structure. For example, developing a `get-out-of-the-way' behavior involves two main steps. The first step is to implement the `get-out' algorithm as an independent module and connecting its output to the mux. The next step is to add the triggering condition of this manuever in the behavior tree.  This ties back to our fifth learned lesson: \ref{l5:easydev}, which promotes faster development of future capabilities. Both convoy mode and get-out-of-the-way behaviors enable one-to-many control and automatic coordination, which significantly reduce the operator's burden when managing multiple robots. \ref{l4:adapiface}




\begin{figure}[b!]
    \centering
    \includegraphics[width = 0.9\linewidth]{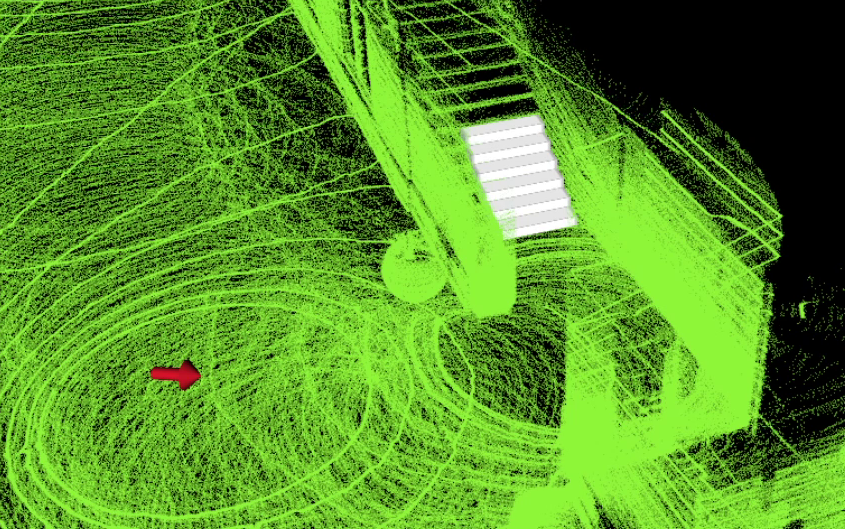}
    \caption{Detected staircases depicted using white markers in the 3D map}
    \label{fig:staircase}
\end{figure}
\subsection{System-wide services}
While our proposed architecture has many definite data channels for passing information between modules, certain components are declared as a system-wide service from which any robot can utilize their output. These services fall into one of two categories: base-level knowledge or query-able services. In our architecture, base-level knowledge incorporates both SLAM and perception, whose outputs are considered requisite for autonomy. For example, a staircase detection algorithm~\cite{sriganesh2023fast} runs on all robots to automatically inform the operator about staircases in the environment. Figure \ref{fig:staircase} shows detected staircases as markers in the RViz map. A query-able service, however, is complex function or behavior, such as a global planner, which may be required by multiple different modules of the system. In both cases, restricting access to the data channel either requires inefficient passing of information through multiple modules, or a module accepting the entirety of the output while it is not needed. Instead, we embrace direct access to these specific system-wide services in which any module can obtain their outputs or call their functions as necessary.
This design supports modularity, allowing independent implementation of algorithms for different platforms that adhere to a standardized data format \ref{l5:easydev}.


\section{Conclusion}
In this paper, we present a system architecture for multi-robot operations. Drawing on lessons learned from our experience in the DARPA SubT challenge, and real-world deployments of multi-robot teams, our approach prioritizes operator control, interoperability, control flow management, adaptive interfaces, and modular design for rapid development. We outline how the key components, such as behavior trees, the mux, and the command interface, work in concert to provide flexible autonomy levels, streamline coordination of heterogeneous robots, and reduce the operator's cognitive workload.

The proposed architecture, with its emphasis on usability and adaptability, holds significant promise for advancing effective robot deployment in challenging, unstructured environments. By carefully considering the human-robot interaction and the practical realities of system failures, our work contributes to the development of more capable and reliable autonomy solutions. In the future, we plan to present the field reports from deploying and testing this architecture on real heterogeneous multi-robot systems. Future investigations will also explore scalability to larger teams and further refinement of adaptive interface elements.

\section*{References}
\printbibliography[heading=none]

@book{Toth2014, author = {Toth, Paolo and Vigo, Daniele}, title = {Vehicle Routing}, publisher = {Society for Industrial and Applied Mathematics}, year = {2014}, address = {Philadelphia, PA}}

@ARTICLE{Nahavandi2022,  author={Nahavandi, Saeid and Mohamed, Shady and Hossain, Ibrahim and Nahavandi, Darius and Salaken, Syed Moshfeq and Rokonuzzaman, Mohammad and Ayoub, Rachael and Smith, Robin},  journal={IEEE Access},   title={Autonomous Convoying: A Survey on Current Research and Development},   year={2022},  volume={10},  number={},  pages={13663-13683}}

@article{Travers2022, author = {Scherer, Sebastian and others}, journal = {Field Robotics Journal}, title = {Resilient and Modular Subterranean Exploration with a Team of Roving and Flying Robots}, pages = {678-734}, month = {May}, year = {2022}}

@article{tatum2020,
  title={Communications coverage in unknown underground environments},
  author={Tatum, Michael},
  journal={Masters’s thesis, The Robotics Institute, Carnegie Mellon University, USA},
  year={2020}
}

@article{norlab_final,
author = {Bayer, Jan and Čížek, Petr and Faigl, Jan},
year = {2023},
month = {01},
pages = {266-300},
title = {Autonomous Multi-robot Exploration with Ground Vehicles in DARPA Subterranean Challenge Finals},
volume = {3},
journal = {Field Robotics}
}

@article{norlab,
  author       = {Tom{\'{a}}s Roucek and
                  Martin Pecka and
                  Petr C{\'{\i}}zek and
                  Tom{\'{a}}s Petr{\'{\i}}cek and
                  Jan Bayer and
                  Vojtech Salansk{\'{y}} and
                  Teymur Azayev and
                  Daniel Hert and
                  Matej Petrl{\'{\i}}k and
                  Tom{\'{a}}s B{\'{a}}ca and
                  Vojtech Spurn{\'{y}} and
                  V{\'{\i}}t Kr{\'{a}}tk{\'{y}} and
                  Pavel Petr{\'{a}}cek and
                  Dominic Baril and
                  Maxime Vaidis and
                  Vladim{\'{\i}}r Kubelka and
                  Fran{\c{c}}ois Pomerleau and
                  Jan Faigl and
                  Karel Zimmermann and
                  Martin Saska and
                  Tom{\'{a}}s Svoboda and
                  Tom{\'{a}}s Krajn{\'{\i}}k},
  title        = {System for multi-robotic exploration of underground environments {CTU-CRAS-NORLAB}
                  in the {DARPA} Subterranean Challenge},
  journal      = {CoRR},
  volume       = {abs/2110.05911},
  year         = {2021},
  bibsource    = {dblp computer science bibliography, https://dblp.org}
}

@article{csiro,
  author       = {Ali Agha and
                  Kyohei Otsu and
                  Benjamin Morrell and
                  David D. Fan and
                  Rohan Thakker and
                  Angel Santamaria{-}Navarro and
                  Sung{-}Kyun Kim and
                  Amanda Bouman and
                  Xianmei Lei and
                  Jeffrey A. Edlund and
                  Muhammad Fadhil Ginting and
                  Kamak Ebadi and
                  Matthew Anderson and
                  Torkom Pailevanian and
                  Edward Terry and
                  Michael T. Wolf and
                  Andrea Tagliabue and
                  Tiago Stegun Vaquero and
                  Matteo Palieri and
                  Scott Tepsuporn and
                  Yun Chang and
                  Arash Kalantari and
                  Fernando Chavez and
                  Brett Thomas Lopez and
                  Nobuhiro Funabiki and
                  Gregory Miles and
                  Thomas Touma and
                  Alessandro Buscicchio and
                  Jesus Tordesillas and
                  Nikhilesh Alatur and
                  Jeremy Nash and
                  William Walsh and
                  Sunggoo Jung and
                  Hanseob Lee and
                  Christoforos Kanellakis and
                  John Mayo and
                  Scott Harper and
                  Marcel Kaufmann and
                  Anushri Dixit and
                  Gustavo Correa and
                  Carlyn Lee and
                  Jay Gao and
                  Gene Merewether and
                  Jairo Maldonado{-}Contreras and
                  Gautam Salhotra and
                  Ma{\'{\i}}ra Saboia da Silva and
                  Benjamin Ramtoula and
                  Yuki Kubo and
                  Seyed Abolfazl Fakoorian and
                  Alexander Hatteland and
                  Taeyeon Kim and
                  Tara Bartlett and
                  Alex Stephens and
                  Leon Kim and
                  Chuck Bergh and
                  Eric Heiden and
                  Thomas Lew and
                  Abhishek Cauligi and
                  Tristan Heywood and
                  Andrew Kramer and
                  Henry A. Leopold and
                  Hyungho Chris Choi and
                  Shreyansh Daftry and
                  Olivier Toupet and
                  Inhwan Wee and
                  Abhishek Thakur and
                  Micah Feras and
                  Giovanni Beltrame and
                  George Nikolakopoulos and
                  David Hyunchul Shim and
                  Luca Carlone and
                  Joel Burdick},
  title        = {NeBula: Quest for Robotic Autonomy in Challenging Environments; {TEAM}
                  CoSTAR at the {DARPA} Subterranean Challenge},
  journal      = {CoRR},
  volume       = {abs/2103.11470},
  year         = {2021},
  bibsource    = {dblp computer science bibliography, https://dblp.org}
}

@article{costar,
  author       = {Ali Agha and
                  Kyohei Otsu and
                  Benjamin Morrell and
                  David D. Fan and
                  Rohan Thakker and
                  Angel Santamaria{-}Navarro and
                  Sung{-}Kyun Kim and
                  Amanda Bouman and
                  Xianmei Lei and
                  Jeffrey A. Edlund and
                  Muhammad Fadhil Ginting and
                  Kamak Ebadi and
                  Matthew Anderson and
                  Torkom Pailevanian and
                  Edward Terry and
                  Michael T. Wolf and
                  Andrea Tagliabue and
                  Tiago Stegun Vaquero and
                  Matteo Palieri and
                  Scott Tepsuporn and
                  Yun Chang and
                  Arash Kalantari and
                  Fernando Chavez and
                  Brett Thomas Lopez and
                  Nobuhiro Funabiki and
                  Gregory Miles and
                  Thomas Touma and
                  Alessandro Buscicchio and
                  Jesus Tordesillas and
                  Nikhilesh Alatur and
                  Jeremy Nash and
                  William Walsh and
                  Sunggoo Jung and
                  Hanseob Lee and
                  Christoforos Kanellakis and
                  John Mayo and
                  Scott Harper and
                  Marcel Kaufmann and
                  Anushri Dixit and
                  Gustavo Correa and
                  Carlyn Lee and
                  Jay Gao and
                  Gene Merewether and
                  Jairo Maldonado{-}Contreras and
                  Gautam Salhotra and
                  Ma{\'{\i}}ra Saboia da Silva and
                  Benjamin Ramtoula and
                  Yuki Kubo and
                  Seyed Abolfazl Fakoorian and
                  Alexander Hatteland and
                  Taeyeon Kim and
                  Tara Bartlett and
                  Alex Stephens and
                  Leon Kim and
                  Chuck Bergh and
                  Eric Heiden and
                  Thomas Lew and
                  Abhishek Cauligi and
                  Tristan Heywood and
                  Andrew Kramer and
                  Henry A. Leopold and
                  Hyungho Chris Choi and
                  Shreyansh Daftry and
                  Olivier Toupet and
                  Inhwan Wee and
                  Abhishek Thakur and
                  Micah Feras and
                  Giovanni Beltrame and
                  George Nikolakopoulos and
                  David Hyunchul Shim and
                  Luca Carlone and
                  Joel Burdick},
  title        = {NeBula: Quest for Robotic Autonomy in Challenging Environments; {TEAM}
                  CoSTAR at the {DARPA} Subterranean Challenge},
  journal      = {CoRR},
  volume       = {abs/2103.11470},
  year         = {2021},
  bibsource    = {dblp computer science bibliography, https://dblp.org}
}

@inproceedings{zhao2021super,
  title={Super odometry: IMU-centric LiDAR-visual-inertial estimator for challenging environments},
  author={Zhao, Shibo and Zhang, Hengrui and Wang, Peng and Nogueira, Lucas and Scherer, Sebastian},
  booktitle={2021 IEEE/RSJ International Conference on Intelligent Robots and Systems (IROS)},
  pages={8729--8736},
  year={2021},
  organization={IEEE}
}

@inproceedings{yang2022far,
  title={FAR planner: Fast, attemptable route planner using dynamic visibility update},
  author={Yang, Fan and Cao, Chao and Zhu, Hongbiao and Oh, Jean and Zhang, Ji},
  booktitle={2022 ieee/rsj international conference on intelligent robots and systems (iros)},
  pages={9--16},
  year={2022},
  organization={IEEE}
}

@inproceedings{bouman2020autonomous,
  title={Autonomous spot: Long-range autonomous exploration of extreme environments with legged locomotion},
  author={Bouman, Amanda and Ginting, Muhammad Fadhil and Alatur, Nikhilesh and Palieri, Matteo and Fan, David D and Touma, Thomas and Pailevanian, Torkom and Kim, Sung-Kyun and Otsu, Kyohei and Burdick, Joel and others},
  booktitle={2020 IEEE/RSJ International Conference on Intelligent Robots and Systems (IROS)},
  pages={2518--2525},
  year={2020},
  organization={IEEE}
}

@article{palieri2021corrections,
  title={Corrections to “LOCUS: A Multi-Sensor Lidar-Centric Solution for High-Precision Odometry and 3D Mapping in Real-Time”[Apr 21 421-428]},
  author={Palieri, Matteo and Morrell, Benjamin and Thakur, Abhishek and Ebadi, Kamak and Nash, Jeremy and Chatterjee, Arghya and Kanellakis, Christoforos and Carlone, Luca and Guaragnella, Cataldo and Agha-mohammadi, Ali-akbar},
  journal={IEEE Robotics and Automation Letters},
  volume={6},
  number={2},
  pages={3760--3760},
  year={2021},
  publisher={IEEE}
}

@article{darpa_tim,
  title={Into the robotic depths: analysis and insights from the {DARPA} subterranean challenge},
  author={Chung, Timothy H and Orekhov, Viktor and Maio, Angela},
  journal={Annual Review of Control, Robotics, and Autonomous Systems},
  volume={6},
  pages={477--502},
  year={2023},
  publisher={Annual Reviews}
}

@article{heger2006sliding,
  title={Sliding autonomy for complex coordinated multi-robot tasks: Analysis \& experiments},
  author={Heger, Frederik W and Singh, Sanjiv},
  year={2006},
  publisher={Carnegie Mellon University}
}

@article{dias2008sliding,
  title={Sliding autonomy for peer-to-peer human-robot teams},
  author={Dias, M Bernardine and Kannan, Balajee and Browning, Brett and Jones, Gil and Argall, Brenna and Dias, M Freddie and Zinck, Marc and Veloso, Manuela M and Stentz, Anthony},
  year={2008},
  publisher={Carnegie Mellon University}
}

@article{bt_survey,
  title={A survey of behavior trees in robotics and ai},
  author={Iovino, Matteo and Scukins, Edvards and Styrud, Jonathan and {\"O}gren, Petter and Smith, Christian},
  journal={Robotics and Autonomous Systems},
  volume={154},
  pages={104096},
  year={2022},
  publisher={Elsevier}
}

@article{bagree2023distributed,
  title={Distributed Optimal Control Framework for High-Speed Convoys: Theory and Hardware Results},
  author={Bagree, Namya and Noren, Charles and Singh, Damanpreet and Travers, Matthew and Vundurthy, Bhaskar},
  journal={IFAC-PapersOnLine},
  volume={56},
  number={2},
  pages={2127--2133},
  year={2023},
  publisher={Elsevier}
}

@inproceedings{whitman2018snake,
  title={Snake robot urban search after the 2017 mexico city earthquake},
  author={Whitman, Julian and Zevallos, Nico and Travers, Matt and Choset, Howie},
  booktitle={2018 IEEE international symposium on safety, security, and rescue robotics (SSRR)},
  pages={1--6},
  year={2018},
  organization={IEEE}
}

@article{fukushimadaiichi,
  title={The use of robots to respond to nuclear accidents: Applying the lessons of the past to the fukushima daiichi nuclear power station},
  author={Yokokohji, Yasuyoshi},
  journal={Annual Review of Control, Robotics, and Autonomous Systems},
  volume={4},
  pages={681--710},
  year={2021},
  publisher={Annual Reviews}
}

@book{colledanchise2018behavior,
  title={Behavior trees in robotics and AI: An introduction},
  author={Colledanchise, Michele and {\"O}gren, Petter},
  year={2018},
  publisher={CRC Press}
}

@inproceedings{baker2004campaign,
  title={A campaign in autonomous mine mapping},
  author={Baker, Christopher and Morris, Aaron and Ferguson, David and Thayer, Scott and Whittaker, Chuck and Omohundro, Zachary and Reverte, Carlos and Whittaker, William and Hahnel, D and Thrun, Sebastian},
  booktitle={IEEE International Conference on Robotics and Automation, 2004. Proceedings. ICRA'04. 2004},
  volume={2},
  year={2004},
  organization={IEEE}
}

@article{thrun2004autonomous,
  title={Autonomous exploration and mapping of abandoned mines},
  author={Thrun, Sebastian and Thayer, Scott and Whittaker, William and Baker, Christopher and Burgard, Wolfram and Ferguson, David and Hahnel, Dirk and Montemerlo, D and Morris, Aaron and Omohundro, Zachary and others},
  journal={IEEE Robotics \& Automation Magazine},
  volume={11},
  number={4},
  pages={79--91},
  year={2004},
  publisher={IEEE}
}

@article{nagatani2013emergency,
  title={Emergency response to the nuclear accident at the Fukushima Daiichi Nuclear Power Plants using mobile rescue robots},
  author={Nagatani, Keiji and Kiribayashi, Seiga and Okada, Yoshito and Otake, Kazuki and Yoshida, Kazuya and Tadokoro, Satoshi and Nishimura, Takeshi and Yoshida, Tomoaki and Koyanagi, Eiji and Fukushima, Mineo and others},
  journal={Journal of Field Robotics},
  volume={30},
  number={1},
  pages={44--63},
  year={2013},
  publisher={Wiley Online Library}
}

@inproceedings{agha2019robotic,
  title={Robotic exploration of planetary subsurface voids in search for life},
  author={Agha, Ali and Mitchell, KL and Boston, PJ},
  booktitle={AGU Fall Meeting Abstracts},
  volume={2019},
  pages={P41C--3463},
  year={2019}
}

@inproceedings{liu2016ssd,
  title={Ssd: Single shot multibox detector},
  author={Liu, Wei and Anguelov, Dragomir and Erhan, Dumitru and Szegedy, Christian and Reed, Scott and Fu, Cheng-Yang and Berg, Alexander C},
  booktitle={Computer Vision--ECCV 2016: 14th European Conference, Amsterdam, The Netherlands, October 11--14, 2016, Proceedings, Part I 14},
  pages={21--37},
  year={2016},
  organization={Springer}
}

@inproceedings{szegedy2016rethinking,
  title={Rethinking the inception architecture for computer vision},
  author={Szegedy, Christian and Vanhoucke, Vincent and Ioffe, Sergey and Shlens, Jon and Wojna, Zbigniew},
  booktitle={Proceedings of the IEEE conference on computer vision and pattern recognition},
  pages={2818--2826},
  year={2016}
}

@article{delmerico2019current,
  title={The current state and future outlook of rescue robotics},
  author={Delmerico, Jeffrey and Mintchev, Stefano and Giusti, Alessandro and Gromov, Boris and Melo, Kamilo and Horvat, Tomislav and Cadena, Cesar and Hutter, Marco and Ijspeert, Auke and Floreano, Dario and others},
  journal={Journal of Field Robotics},
  volume={36},
  number={7},
  pages={1171--1191},
  year={2019},
  publisher={Wiley Online Library}
}

@book{murphy2016disaster,
  title={Disaster robotics},
  author={Murphy, Robin R and Tadokoro, Satoshi and Kleiner, Alexander},
  year={2016},
  publisher={Springer}
}

@article{amigoni2017multirobot,
  title={Multirobot exploration of communication-restricted environments: A survey},
  author={Amigoni, Francesco and Banfi, Jacopo and Basilico, Nicola},
  journal={IEEE Intelligent Systems},
  volume={32},
  number={6},
  pages={48--57},
  year={2017},
  publisher={IEEE}
}

@inproceedings{quigley2009ros,
  title={ROS: an open-source Robot Operating System},
  author={Quigley, Morgan and Conley, Ken and Gerkey, Brian and Faust, Josh and Foote, Tully and Leibs, Jeremy and Wheeler, Rob and Ng, Andrew Y and others},
  booktitle={ICRA workshop on open source software},
  volume={3},
  number={3.2},
  pages={5},
  year={2009},
  organization={Kobe, Japan}
}

@inproceedings{sriganesh2023fast,
  title={Fast Staircase Detection and Estimation using 3D Point Clouds with Multi-detection Merging for Heterogeneous Robots},
  author={Sriganesh, Prasanna and Bagree, Namya and Vundurthy, Bhaskar and Travers, Matthew},
  booktitle={2023 IEEE International Conference on Robotics and Automation (ICRA)},
  pages={9253--9259},
  year={2023},
  organization={IEEE}
}

@article{zhang2020falco,
  title={Falco: Fast likelihood-based collision avoidance with extension to human-guided navigation},
  author={Zhang, Ji and Hu, Chen and Chadha, Rushat Gupta and Singh, Sanjiv},
  journal={Journal of Field Robotics},
  volume={37},
  number={8},
  pages={1300--1313},
  year={2020},
  publisher={Wiley Online Library}
}

@article{chen2007human,
  title={Human performance issues and user interface design for teleoperated robots},
  author={Chen, Jessie YC and Haas, Ellen C and Barnes, Michael J},
  journal={IEEE Transactions on Systems, Man, and Cybernetics, Part C (Applications and Reviews)},
  volume={37},
  number={6},
  pages={1231--1245},
  year={2007},
  publisher={IEEE}
}

@incollection{mayne1973differential,
  title={Differential dynamic programming--a unified approach to the optimization of dynamic systems},
  author={Mayne, David Q},
  booktitle={Control and dynamic systems},
  volume={10},
  pages={179--254},
  year={1973},
  publisher={Elsevier}
}
\end{document}